\documentclass[10pt,conference,a4paper]{IEEEtran}
\ifCLASSINFOpdf
\else
\fi
\usepackage{times}
\usepackage{epsfig}
\usepackage{graphicx}
\usepackage{amsmath}
\usepackage{amssymb}
\usepackage[ruled]{algorithm2e}
\usepackage{enumerate}
\usepackage{enumitem}
\usepackage{bm}
\usepackage{booktabs}
\usepackage{mathtools}
\usepackage{algorithmic}

\usepackage[pagebackref=true,breaklinks=true,letterpaper=true,colorlinks,bookmarks=false]{hyperref}

\DeclareMathOperator*{\argmin}{arg\,min}

\begin{document}
\title{A Self-supervised GAN for Unsupervised Few-shot Object Recognition}

\author{Khoi Nguyen and Sinisa Todorovic\\
Oregon State University\\
Corvallis, OR 97330, USA\\
{\tt\small {\{nguyenkh,sinisa\}}@oregonstate.edu}
}


%Telephone: (800) 555--1212, Fax: (888) 555--1212}


\maketitle

\begin{abstract}
This paper addresses unsupervised few-shot object recognition, where all training images are unlabeled, and test images are divided into queries and a few labeled support images per object class of interest. The training and test images do not share object classes.  We extend the vanilla GAN with two loss functions, both aimed at self-supervised learning. The first is a reconstruction loss that enforces the discriminator to reconstruct the probabilistically sampled latent code which has been used for generating the ``fake'' image. The second is a triplet loss that enforces the discriminator to output image encodings that are closer for more similar images. Evaluation, comparisons, and detailed ablation studies are done in the context of few-shot classification. Our approach significantly outperforms the state of the art on the Mini-Imagenet and Tiered-Imagenet datasets.
 \end{abstract}


% \fi
\IEEEpeerreviewmaketitle

\section{Introduction}
\label{sec:intro}
% This paper presents a new deep architecture for unsupervised learning of image representations which are suitable for subsequently addressing a recognition task in a new domain.  With access to only unlabeled data, our goal is to learn a useful representation such that when presented with a new domain with classes that do not appear in the unlabeled dataset, the model can generalize well to the new classes without additional re-training, domain adaptation or transfer learning to the new domain. 

This paper presents a new deep architecture for unsupervised few-shot object recognition. In training, we are given a set of unlabeled images. In testing, we are given a small number $K$ of {\em support images} with labels sampled from $N$ object classes that do not appear in the training set (also referred to as unseen classes). Our goal in testing is to predict the label of a \textit{query image} as one of these $N$ previously unseen classes. A common approach to this $N$-way $K$-shot recognition problem is to take the label of the closest support to the query. Thus, our key challenge is to learn a deep image representation on unlabeled data such that it would in testing generalize well to  unseen classes, so as to enable accurate distance estimation between the query and support images.

This problem is important as it appears in a wide range of applications. For example, we expect that leveraging unlabeled data could help few-shot image classification in domains with  very few labeled images per class (e.g., medical images). Another example is  online tracking of a previously unseen object in a video, initialized with a single bounding box. Unsupervised few-shot object recognition is different from the standard few-shot learning \cite{snell2017prototypical,finn2017model} that has access to a significantly larger set of labeled images, allowing for episodic training \cite{vinyals2016matching}. Episodic training cannot be used in our setting with a few annotations. 

%In general, our problem can be viewed as a special case of semi-supervised learning where the unlabeled and labeled datasets do not share the same classes, and the labeled dataset is too small to allow for standard re-training, domain adaptation, or knowledge transfer from unlabeled to labeled data.

% Out of many applications, in this paper, we evaluate our unsupervised image representation on the following tasks: (I) few-shot  classification, (II) K-nearest neighbor  retrieval, and (III) K-means clustering. In (I), we evaluate accuracy of labeling query images with one of $N$ classes given a small number $K$ of labeled support images per class ($K\le 5$). We follow a common approach to take the label of the closest support to the query. In (II), we evaluate a percentage of the correctly retrieved images among the $K$ nearest neighbors of a given query image. In (III), we evaluate the normalized mutual information of the K-means clustering of our unsupervised deep features representing test images.

\begin{figure*}[t!]
    \centering
    \begin{minipage}[t]{0.77\textwidth}
    \centering
    \includegraphics[scale=0.36]{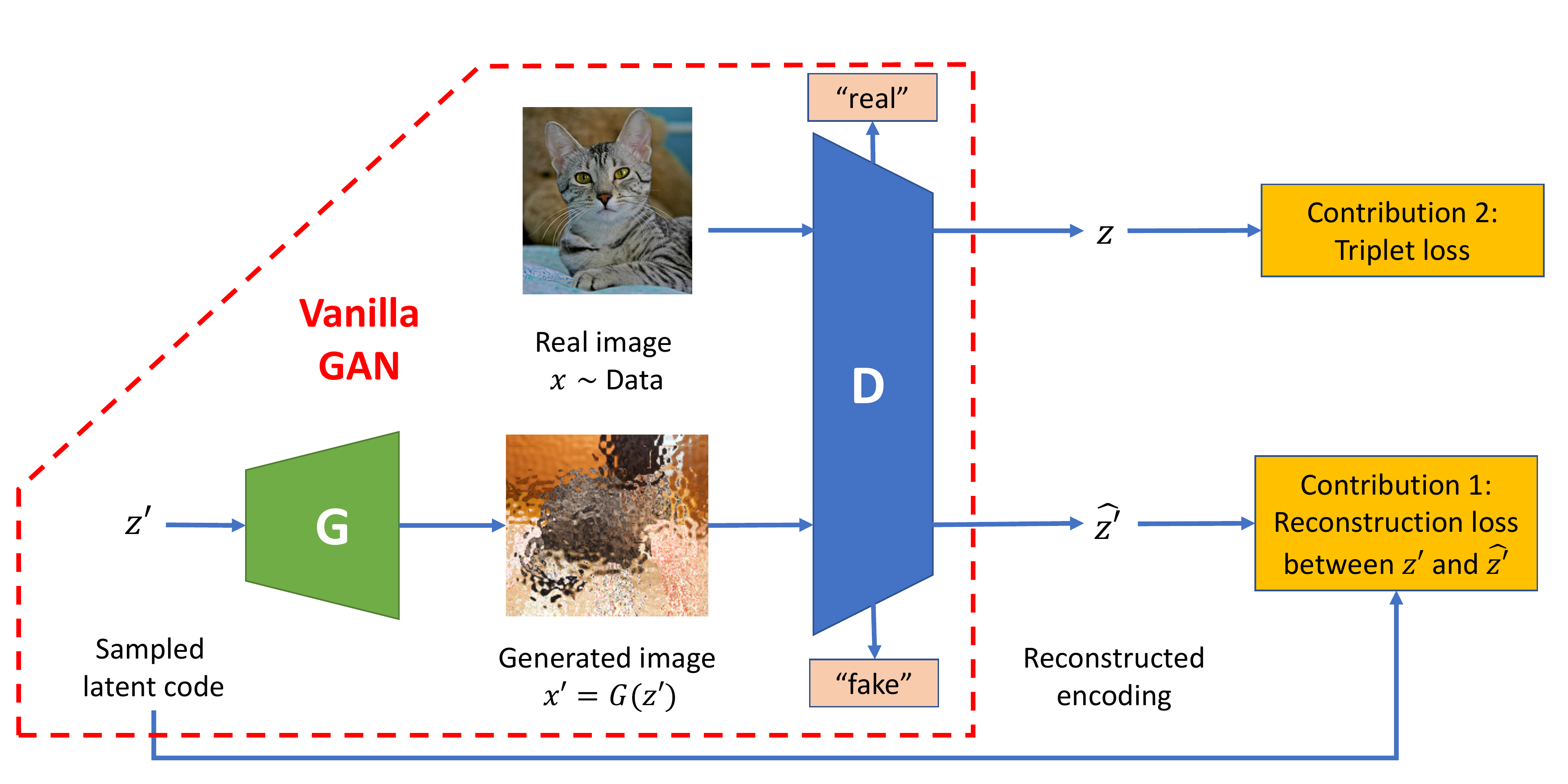}
    \caption{We extend the vanilla GAN to learn an image encoding $z$ on unlabeled data that will be suitable for subsequent few-shot image classification and image retrieval in new domains with very few annotations. Our extension integrates self-supervised and adversarial learning by the means of: (a)  Reconstruction loss so the encoding $\hat{z'}$ of a ``fake'' image is similar to the corresponding randomly sampled code $z'$; and (b) Deep metric learning so the image encodings $z$ are closer for similar ``real'' images than for dissimilar ones.}
    \label{fig:overview}
    \end{minipage}\hfill%
        \begin{minipage}[t]{0.2\textwidth}
        \centering
    \includegraphics[scale=0.3]{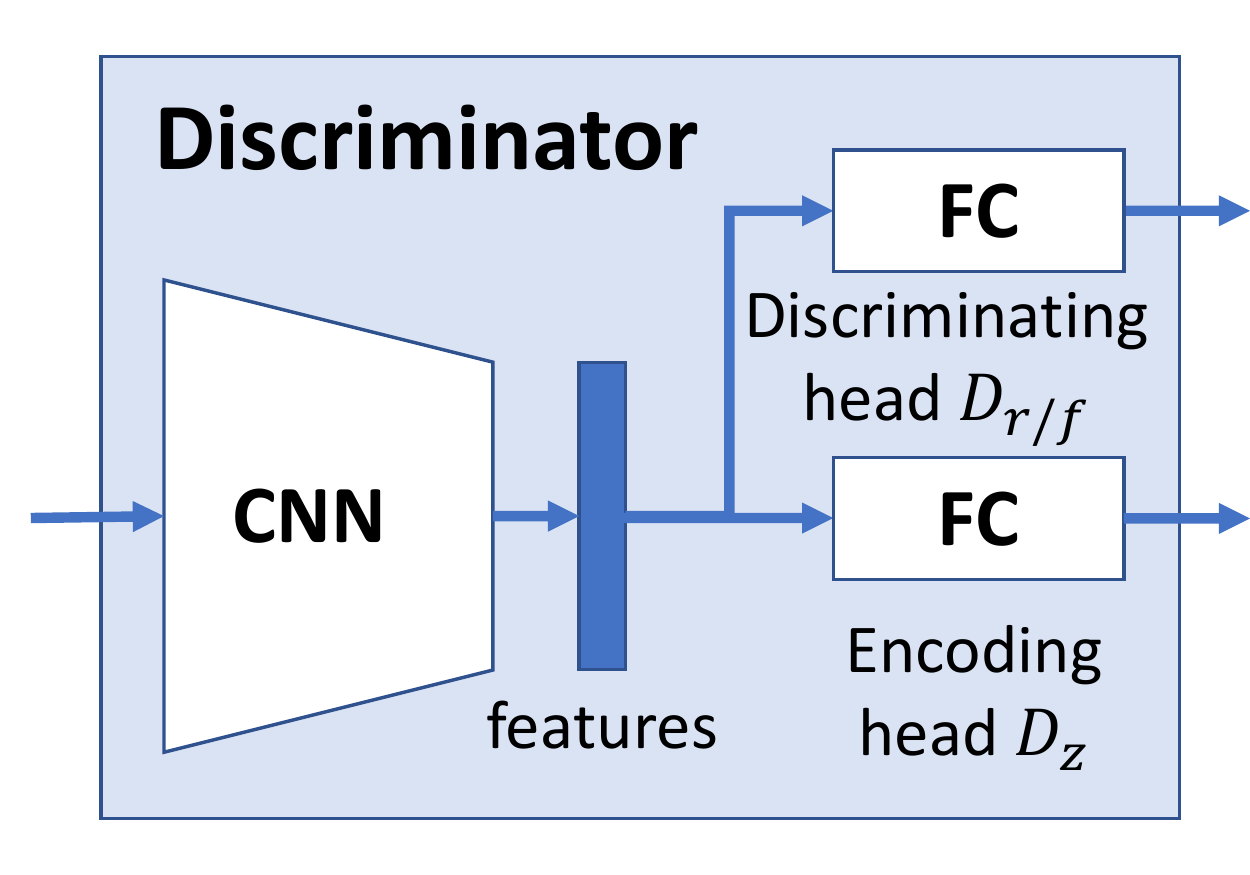}
    \caption{Details of the discriminator from Fig.~\ref{fig:overview} with discriminating head and encoding head.}
    \label{fig:discriminator}
        \end{minipage}
\end{figure*}

There is scant work on unsupervised few-shot classification. Recent work \cite{hsu2018unsupervised,antoniou2019assume,khodadadeh2018unsupervised} first identifies pseudo labels of unlabeled training images, and then uses the standard episodic training \cite{vinyals2016matching} on these pseudo labels. However, performance of these methods is significantly below that of counterpart approaches to supervised few-shot learning. 

% K-nearest-neighbor retrieval and K-means clustering have also been used for evaluation of unsupervised distance-metric learning \cite{iscen2018mining, ye2019unsupervised}. However, this work  seems a misnomer, since they claim unsupervised training and yet they use full supervision while pre-training their models on the ImageNet classes \cite{russakovsky2015imagenet}. 

Motivated by the success of Generative Adversarial Networks (GANs)  \cite{goodfellow2014generative, donahue2016adversarial, miyato2018spectral} to generalize well to new domains, we adopt and extend this framework with two new strategies for self-supervision \cite{SelfSupervision_Survey}.  A GAN is appropriate for our problem since it is a generative model aimed at learning the underlying image prior in an unsupervised manner, rather than discriminative image features which would later be difficult to ``transfer'' to new domains.  As shown in Fig.~\ref{fig:overview}, a GAN consists of a generator and discriminator that are adversarially trained such that the discriminator distinguishes between ``real'' and ``fake'' images, where the latter are produced by the generator from randomly sampled latent codes. We extend this framework by allowing the discriminator not only to predict the ``real'' or ``fake'' origins of the input image but also to output a deep image  feature, which we will use later for unsupervised few-shot classification task. This allows us to augment the standard adversarial learning of the extended GAN with additional self-supervised learning via two loss functions -- reconstruction loss and distance-metric triplet loss.  

{\bf Our first contribution:} By minimizing a reconstruction loss between the randomly sampled code and the discriminator's encoding of the ``fake'' image, we enforce the discriminator to explicitly capture the most relevant characteristics of the random codes that have been used to generate the corresponding ``fake'' images. In this way, the discriminator seeks to extract relevant features from images which happen to be ``fake'' but are guaranteed by the adversarial learning to be good enough to fool the discriminator of their origin. Thus, we use the randomly sampled codes not only for the adversarial learning but also as a ``free'' ground-truth for self-supervised learning  \cite{SelfSupervision_Survey}. From our experiments, the added reconstruction loss gives a significant performance improvement over the vanilla GAN.

\begin{figure}[b]
    \centering
    \includegraphics[scale=0.35]{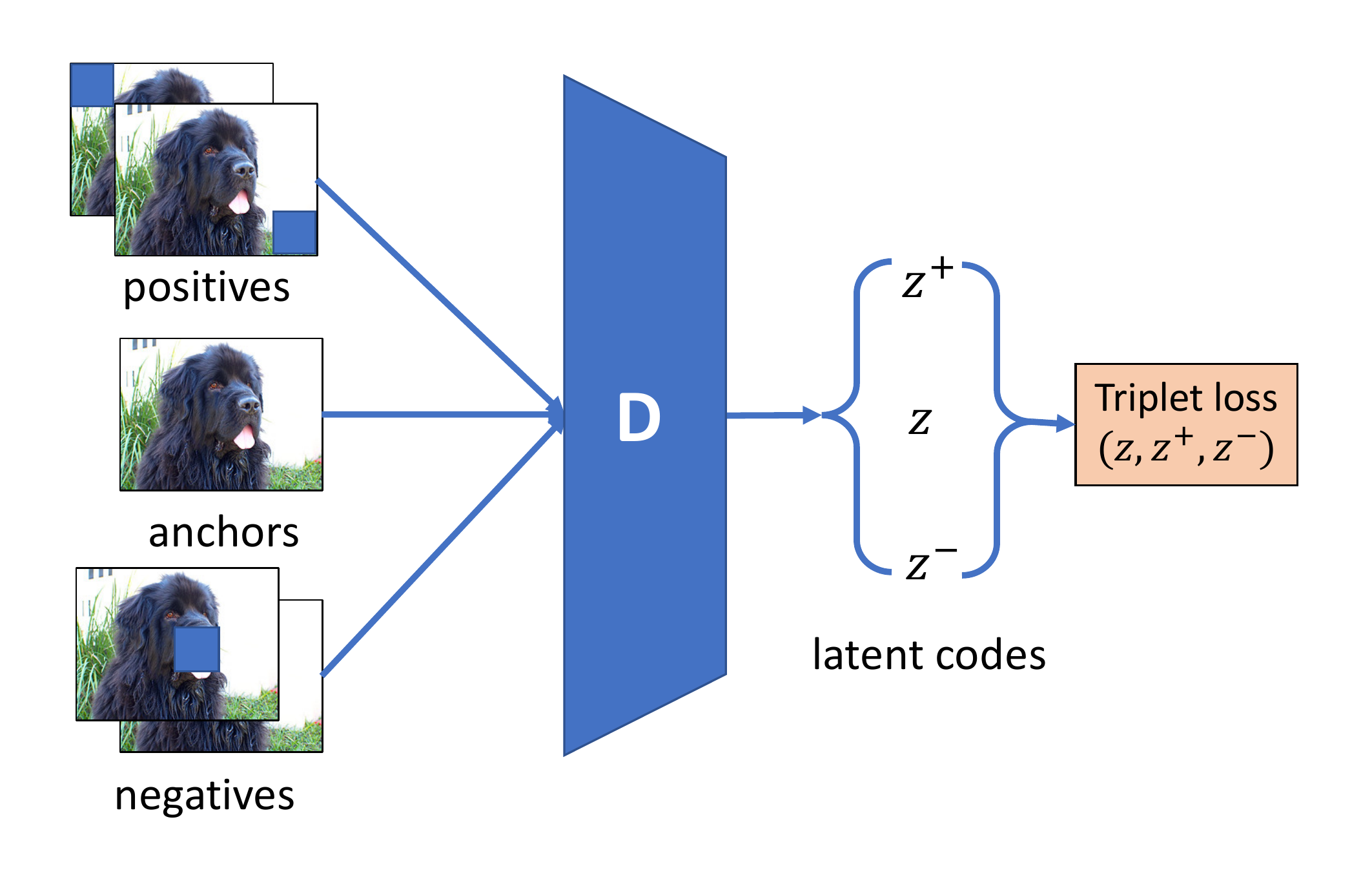}
    \caption{Self-supervised learning of our extended GAN by minimizing the triplet loss of  $\langle$anchor, positive,  negative$\rangle$ images, where the positives and negatives are generated via appropriately masking the anchor. D is the discriminator show in Fig.~\ref{fig:discriminator}}
    \label{fig:stage2}
\end{figure}

{\bf Our second contribution:} As shown in Fig.~\ref{fig:overview}, we specify another type of self-supervised learning for our extended GAN so image encodings at the discriminator's output respect similarity of input images. While in general this additional self-supervision along with adversarial learning is expected to  produce a better GAN model, it is particularly suitable for our subsequent distance-based image retrieval and few-shot image classification. In the lack of labeled data, for distance-metric learning, we resort to data augmentation. We take ``real'' training images and mask them with a patch whose placement controls similarity between the masked and original image, as shown in Fig.~\ref{fig:stage2}. Following the long track of research on object center-bias \cite{borji2015reconciling, henderson1993eye, nuthmann2010object, elazary2008interesting} arguing that human attention usually focuses on an object in the image center,  we declare the masked images more similar to the original if their masking patches fall in the image corners rather than on the image center, as the latter case is more likely to partially occlude the target object. 

We use the above masking procedure to compile a set of image triplets that encode similarity relationships between the images. In a triplet, the anchor is a ``real'' image selected from the training dataset, the positive is a masked version of the anchor with the masking patch in one of the image corners, and the negative is another masked anchor but with the masking patch falling close to the image center. Thus, by construction, we constrain the triplet's similarity relationships relative to the anchor, as our training images are unlabeled. %instead of selecting for the positives and negatives different training images. This avoids errors when two rather distinct images with very close embeddings are declared as the anchor and positive, or conversely when two images of the same class are declared as the anchor and negative. 
Given the set of triplets, we estimate the standard triplet loss for our GAN training. As our results show, the added distance-metric learning further improves the subsequent few-shot classification.

Our two contributions lead to significant performance gains in unsupervised few-shot image classification relative to recent unsupervised approaches  on the Mini-Imagenet \cite{vinyals2016matching,ravi2016optimization} and Tiered-Imagenet \cite{ren2018meta} datasets.

% NEED TO REWRITE to match new structure
In the rest of this paper: Sec.~\ref{sec:related_work} reviews prior work, Sec.~\ref{sec:proposed_model} specifies our approach, and Sec.~\ref{sec:experiments} presents our implementation details and  experimental results. 

\section{Related Work}
\label{sec:related_work}
This section reviews  closely related work.

Methods for \textbf{deep unsupervised learning} can be broadly divided into: deep clustering, self-supervised learning and generative models. Deep clustering iteratively trains a CNN  \cite{caron2018deep,xie2016unsupervised,wu2018unsupervised}. In each iteration, deep feature extracted by the current CNN are clustered to produce pseudo-labels of training images. The pseudo-labels are then used for the standard supervised learning of the next CNN. Self-supervised methods seek to solve an auxiliary task, for which ground-truth can be easily specified   \cite{doersch2015unsupervised,zhang2016colorful,noroozi2016unsupervised,noroozi2017representation,zhang2017split}. Generative models seek to generate new images that look similar to images sampled from the underlying distribution. Recent generative models include (variational) auto-encoders \cite{kingma2013auto,vincent2010stacked}, and GANs   \cite{goodfellow2014generative,donahue2016adversarial}. Our extended GAN belongs to the group of generative models enhanced with  self-supervised learning.

For \textbf{unsupervised few-shot classification}, recent work uses  unlabeled training data and very few labeled  test data that  do not share the same classes. These methods first assign pseudo labels to the training data by either image clustering \cite{hsu2018unsupervised} or treating each training example as having a unique class \cite{antoniou2019assume, khodadadeh2018unsupervised}. Then, they apply the standard fully-supervised few-shot learning (e.g., \cite{finn2017model, snell2017prototypical}) on the pseudo-labeled training data. We differ from these approaches in two ways. First, we do not estimate pseudo-labels, and do not use the common episodic training for fully-supervised few-shot learning, but seek to directly learn the underlying image prior distribution by integrating adversarial and self-supervised learning. Second, we ensure that our image representation respects similarity relationships of images.

% \textbf{Unsupervised distance metric learning} seeks image embeddings that are close for similar images. As training images are unlabeled, these methods typically compute a pseudo-triplet loss by estimating nearest neighbors   \cite{iscen2018mining}, or using data-augmented versions of the anchor as positive images  and randomly picked other images as negatives \cite{ye2019unsupervised}. However, training images of different classes may be nearest neighbors and training images of the same class may be confused as the negatives -- all introducing error in distance metric learning. We also assign positive and negative pseudo-labels to training images, but seek to reduce errors in the pseudo-labeling by constraining that the positive and negative images in a triplet are appropriately masked versions of the anchor image. 

Our problem is related to semi-supervised few-shot learning \cite{ren2018meta, sun2019learning}. These approaches first augment their labeled training set with unlabeled images, then, apply label propagation for knowledge transfer from the labeled to unlabeled set, and finally conduct fully-supervised few-shot learning on all training data.  We cannot use this framework, as our labeled images have different classes from unlabeled ones, and even if they shared the same classes label, propagation from just one labeled example per class would be  very difficult.  Adversarial learning and self-supervised learning has been integrated  for fully supervised few-shot learning \cite{zhang2018metagan, gidaris2019boosting}, but cannot be easily extended to our unsupervised setting. 

Compared with very closely-related work \cite{chen2019self,tran2019self,doersch2017multi}, our approach significantly differs. \cite{chen2018self,tran2019self} use the same GAN but with a rotation-based self-supervision loss which we do not use. \cite{doersch2017multi} uses a completely different triplet loss from ours. To form a triplet, we use the anchor, positive and negative from the very {\bf same image}, whereas \cite{doersch2017multi} uses the anchor and positive from the same image and the negative from a different image.

\section{Our Approach}
\label{sec:proposed_model}

This section first specifies our extended GAN and its adversarial learning, Sec.~\ref{sec:reconstruction} defines the reconstruction loss, Sec.~\ref{sec:distance} presents our masking procedure and triplet loss, and Sec.~\ref{subsec:training} describes our unsupervised training.  

As shown in Figures~\ref{fig:overview} and \ref{fig:discriminator},  our extended GAN consists of the standard generator network $G$ and a discriminator network $D$ which is equipped with image encoding head $D_z$ and the standard discriminating head $D_{\text{r/f}}$. For adversarial training, we probabilistically sample latent codes $z'$ from a prior distribution,  $z' \sim p(z')$, and generate corresponding ``fake'' images $x' = G(z')$. Both ``real'' training images $x$ and ``fake'' images $x'$ are passed to the discriminator for  $D_{\text{r/f}}$ to predict their ``real'' or ``fake'' nature, $D_{\text{r/f}}(x)\in\mathbb{R}$.

For training $D$ and $G$, we use the standard adversarial loss, specified  as%
\begin{eqnarray}
    L_{D}^{\text{adv}} &=&\underset{x \sim p_{\text{data}}(x)}{\mathrm{E}}[\max (0,1-D_{\text{r/f}}(x))]  \notag \\ 
    && + \underset{z' \sim p(z')}{\mathrm{E}}[\max (0,1+D_{\text{r/f}}(G(z')))], \label{eq:hinge_discriminator}\\
    L_{G}^{\text{adv}} &=&-\underset{z' \sim p(z')}{\mathrm{E}}[D_{\text{r/f}}(G(z'))], 
    \label{eq:hinge_generator}
\end{eqnarray} 
%
%\begin{align}
%    L_{D}^{\text{adv}} &=\underset{x %\sim p_{\text{data}}(x)}{\mathrm{E}}[\m%in (0,-1+D_{\text{r/f}}(x))]  \notag \\ %
%    & + \underset{z' \sim %p(z')}{\mathrm{E}}[\min %(0,-1-D_{\text{r/f}}(G(z')))], %\label{eq:hinge_discriminator} 
%    \end{align}
% \begin{equation}   
 %   L_{G}^{\text{adv}} &=-\underset{z' %\sim p(z')}{\mathrm{E}}[D_{\text{r/f}}%(G(z'))], 
 %   \label{eq:hinge_generator}
%\end{equation}
%
%
%
%
where $E$ denotes the expected value and $p_{\text{data}}(x)$ is a prior distribution of unlabeled ``real'' training images $x$. As shown in \cite{lim2017geometric}, optimizing  \eqref{eq:hinge_discriminator} and \eqref{eq:hinge_generator} is equivalent to minimizing the reverse KL divergence.

For sampling latent codes, $z' \sim p(z')$, as in related work \cite{goodfellow2014generative, donahue2016adversarial, miyato2018spectral}, we assume that all elements of $z'$ are i.i.d.. For the $d$-dimensional latent code $z'$, we have studied several definitions of the prior $p(z')$, including the continuous uniform distribution $p(z')=U[-1, 1]^d$,  the corresponding discrete Bernoulli distribution with the equal probability of 0.5 for the binary outcomes ``1'' and ``-1'' of elements of $z'$, and the Gaussian distribution with the mean zero and variance 1, $p(z')=\mathcal{N}(0,1)^d$. All these definitions of $p(z')$ give similar performance in our experiments. Sampling from the uniform distribution is justified, because many object classes appearing in training images are likely to share the same latent features, so the presence of these features encoded in $z$ and $z'$ is likely to follow the uniform distribution.

\subsection{Reconstruction Loss}
\label{sec:reconstruction}

We extend the above specified adversarial learning of our GAN by enforcing the discriminator's encoding head $D_z$ to reconstruct randomly sampled latent codes $z'\sim p(z')$ from the corresponding ``fake'' images, $\hat{z'}=D_z(G(z'))\in\mathbb{R}^d$, as illustrated in Fig.~\ref{fig:overview}.  Thus, we use $z'$ as a ``free'' label for the additional self-supervised training of $D$ and $G$ along with the adversarial learning. Minimizing a reconstruction loss between $z'$ and $\hat{z'}$ enforces $D_z$ to be trained on ``free'' labels of the corresponding ``fake'' images -- the knowledge that will be later transferred for encoding ``real'' images by $D_z$ when ``fake'' images become good enough to ``fool'' $D_{\text{r/f}}$.

A difference between $z'$ and its reconstruction $\hat{z'}=D_z(G(z'))$ is penalized with the mean-squared error (MSE) reconstruction loss:
\begin{align}
L_{D}^{\text{mse}} = L_{G}^{\text{mse}} = \|\hat{z}' - z'\|^2_2,
\label{eq:mse_discriminator}    
\end{align}
when elements of $z'$ are sampled from the uniform or Gaussian distribution, $p(z')=U[-1, 1]^d$ or $p(z')=\mathcal{N}(0,1)^d$, or with the binary cross-entropy (BCE) reconstruction loss:
\begin{align}
L_{D}^{\text{bce}}  = L_{G}^{\text{bce}} =& -\frac{1}{d}\sum_{m=1}^d[\frac{1+z_m'}{2} \cdot \log{\sigma(\hat{z'}_m)}  \notag \\
& +(1-\frac{1+z_m'}{2})\cdot\log(1 - \sigma(\hat{z'}_m))],
% \end{split}
\label{eq:bce_discriminator} 
\end{align}
when elements of $z'$, $z_m'\in\{-1,1\}$, are sampled from the discrete Bernoulli distribution, $p(z_m'=\pm 1)=0.5$; $\sigma(\cdot)$ denotes the sigmoid function.

\subsection{Image Masking and Triplet Loss}
\label{sec:distance}

We additionally train the discriminator to output image representations that respect image similarity relationships, such that $z_i=D_z(x_i)$ and $z_j=D_z(x_j)$ are closer than $z_i$ and $z_k=D_z(x_k)$ when images $x_i$ and $x_j$ are more similar than $x_i$ and $x_k$. 

As ground-truth similarity relationships are not provided for our training images, we resort to data augmentation using the following masking procedure. Each training image generates a set of its masked versions by exhaustively placing a masking patch on a regular grid covering the image. The masking patch brightness is uniform and equal to the pixel average of the original image. We have experimented with various patch sizes and shapes, and various grids. A good trade-off between complexity and performance for $64\times 64$ training images that we use is obtained for the $16 \times 16$ square patch and $4 \times 4$ regular grid.

Given the training dataset and its masked images, we compile a set of image triplets $\langle$anchor, positive, negative$\rangle$. The anchor $z=D_z(x)$ is an image from the training set. The positive is one of four masked versions of the anchor image $\{z^+_i=D_z(x^+_i):i=1,\dots,4\}$ whose masking patch falls in the image corner -- namely, the top-left, top-right, bottom-left, and bottom-right corner.  The negative is one of the remaining masked versions of the anchor image $\{z^-_j=D_z(x^-_j):j=1,2,\dots\}$ whose masking patch falls on central locations on the grid. In the positives, the relatively small masking patch masks very little to no foreground of the anchor image. On the other hand, in the negatives, the masking patch is very likely to partially occlude foreground of the anchor image. Therefore, by construction, we constrain image similarity relationships to pertain to the image's foreground, such that the positives should be closer to the anchor than the negatives. 

We use the set of triplets $\{\langle z, z_i^+, z_j^- \rangle\}$ to estimate the following triplet loss for the additional distance-metric learning of our GAN:%
\begin{align}
\begin{split}
L_{D}^{\text{triplet}} = \max[0, \max_i \Delta(z, z^+_i) - \min_j \Delta(z, z^-_j) + \rho].
\end{split}
\label{eq:triplet} 
\end{align}
where $\rho\ge 0$ is a distance margin, and $\Delta$ is a distance function. In this paper, we use the common cosine distance:%
\begin{equation}
 \Delta(z, z') = 1 - \frac{z ^\top z'}{\|z\|_2 \cdot \|z'\|_2}.
\label{eq:distance}
 \end{equation}

% \textbf{Discussion:}
% Our triplet loss based on masking strategy is related to random erasing data augmentation \cite{zhong2017random}. However, masking strategy in \cite{zhong2017random} is used for data augmentation to robustly train deep learning model for supervised task such classification, person re-ID or object detection under occlusion scenario. Our approach employ masking strategy in different way, i.e. by designing triplet loss to encourage encoding feature representing salient object, and for different problem setting, i.e. unsupervised training. Furthermore, there are plenty of work about object center-bias \cite{borji2015reconciling, henderson1993eye, nuthmann2010object, elazary2008interesting} explaining why human attention usually focuses on the target object in the middle of their views. That reflects the images taken by human usually contain the salient objects in the middle. This inspires us to design our novel triplet loss based on masking strategy.

\begin{algorithm}[ht]
{\footnotesize
\caption{Our Unsupervised Training}
\label{algo:1} 
\begin{algorithmic}
\STATE {\bf Input} = $\{T_1, T_2, \mathcal{T}, \beta, \gamma, \lambda\}$= numbers of training iterations, and 3 non-negative hyper parameters.
\STATE \COMMENT {First stage $(1)$\quad}
    \FOR {$t_1=1,\dots,T_1$ } 
      	\STATE Sample $z' \sim p(z')$ and reconstruct $\hat{z'} = D_z(G(z'))$;
        \STATE Compute: $L_G^{\text{adv}}$ as in \eqref{eq:hinge_generator}, and $L_G^{\text{mse}}$ as in \eqref{eq:mse_discriminator} or $L_G^{\text{bce}}$ as in \eqref{eq:bce_discriminator};
        \STATE Update $G$ by back-propagating the total loss: \\ \quad $L_G^{(1)} = L_G^{\text{adv}} + \beta L_G^{\text{bce}}$ or  $L_G^{(1)} = L_G^{\text{adv}} + \beta L_G^{\text{mse}}$.
        \FOR {$\tau=1,\dots,\mathcal{T}$}
              	\STATE Sample $z' \sim p(z')$ and reconstruct $\hat{z'} = D_z(G(z'))$;
      	     	\STATE Sample a real image $x \sim p_{\text{data}}(x)$;
      	    	\STATE Compute:  $L_D^{\text{adv}}$ as in \eqref{eq:hinge_discriminator};
      	    	\STATE Compute: $L_G^{\text{mse}}$ as in \eqref{eq:mse_discriminator} or $L_D^{\text{bce}}$ as in \eqref{eq:bce_discriminator};
      	     	\STATE Update $D^{(1)}$ by back-propagating the total loss:\\  \quad $L_D^{(1)} = L_D^{\text{adv}} + \gamma L_D^{\text{bce}}$ or $L_D^{(1)} = L_D^{\text{adv}} + \gamma L_D^{\text{mse}}$.
      	\ENDFOR 
    \ENDFOR
    \STATE \COMMENT {Second stage $(2)$\quad}
        \FOR {$t_2=1,\dots,T_2$}
        \STATE Sample a real image $x \sim p_{\text{data}}(x)$ and take it as the anchor;
        \STATE Generate the masked positives and negatives of the anchor;
        \STATE Form the set of triplets for the anchor;
    \STATE Compute $L_D^{\text{triplet}}$ as in
    \eqref{eq:triplet};
    \STATE Update $D^{(2)}$ by back-propagating the total loss:\\ \quad  $L_D^{(2)} = L_D^{\text{triplet}} + \lambda \|D_z^{(1)}(x) - D_z^{(2)}(x)\|_2^2$. 
    
    	\ENDFOR
    	\STATE Take $D^{(2)}$ as the learned discriminator $D$.
\end{algorithmic}
}
\end{algorithm}

\subsection{Our Unsupervised Training}
\label{subsec:training}
Alg.~\ref{algo:1} summarizes our unsupervised training that integrates  adversarial learning with distance-metric learning and latent-code reconstruction. For easier training of $D$ and $G$, we divide learning in two stages. First, we perform  adversarial learning constrained with the latent-code reconstruction regularization over $t_1=1,\dots,T_1$ iterations. In every iteration, $G$ is optimized once and $D$ is optimized multiple times over $\tau=1,\dots,\mathcal{T}$ iterations ($\mathcal{T}=3$). After convergence of the first stage ($T_1=50,000$), the resulting discriminator is saved as $D^{(1)}$. In the second training stage, we continue with distance-metric learning in $t_2=1,\dots,T_2$ iterations ($T_2=20,000$), while simultaneously regularizing that the discriminator updates do not significantly deviate from the previously learned $D^{(1)}$.

% For K-nearest-neighbor retrieval, we are given a query image $x_q$ and test images $x_i$ whose ground-truth classes are used only for evaluation. We first compute deep representations $z_q=D_z(x_q)$ and $z_i=D_z(x_i)$, and then estimate $K$ nearest neighbors among the $z_i$'s using a distance function (e.g.,  \eqref{eq:distance}). Our retrieval results are expressed as a percentage of the retrieved images that have the same ground-truth class as the query $R@K$. Approaches to distance-metric learning have the same setup for evaluation on K-nearest-neighbor image retrieval \cite{iscen2018mining, ye2019unsupervised}.

% For K-means clustering, we are given  test images whose ground-truth classes are used only for evaluation. Our unsupervised representations of the test images are grouped into $K$ clusters by the K-means algorithm. We set $K$ to be equal to the true number of classes in the test dataset. We evaluate K-means clustering with the normalized mutual information (NMI). For the obtained clusters, $\hat{\Omega} = \{\hat{\omega}_1, \ldots, \hat{\omega}_K\}$ and the ground-truth image clustering by their classes, $\Omega = \{\omega_1, \ldots, \omega_K\}$, the NMI is defined as $\text{NMI}(\hat{\Omega}, \Omega) = \frac{2I(\hat{\Omega}, \Omega)}{H(\hat{\Omega}) + H(\Omega)}$,
%  where $I(\cdot)$ denotes the mutual information, and $H(\cdot)$ is entropy.
% Approaches to distance-metric learning have the same setup for evaluation on K-means image clustering \cite{iscen2018mining, ye2019unsupervised}.

\section{Experiments}
\label{sec:experiments}
% \textbf{Datasets:} We evaluate our approach on the two common few-shot learning datasets: Mini-Imagenet \cite{vinyals2016matching,ravi2016optimization} and Tiered-Imagenet \cite{ren2018meta}. \textit{Mini-Imagenet} contains 100 randomly chosen classes from ILSVRC-2012 \cite{russakovsky2015imagenet}. We split these 100 classes into 64, 16 and 20 classes for meta-training, meta-validation, and meta-testing respectively. Each class contains 600 images of size $84 \times 84$. 

% \textit{Tiered-Imagenet} is a larger subset of ILSVRC-2012 \cite{russakovsky2015imagenet}, consists of 608 classes grouped into 34 high-level categories. These are divided into 20, 6 and 8 categories for meta-training, meta-validation, for meta-testing. This corresponds to 351, 97 and 160 classes for meta-training, meta-validation, and meta-testing respectively. This dataset aims to minimize the semantic similarity between the splits as in Mini-Imagenet. All images are also of size $84 \times 84$. 

% We are the first to report results of unsupervised few-shot recognition on the Tiered-Imagenet dataset.

% For our unsupervised few-shot problem, we ignore all ground-truth labeling information in the training and validation sets, and only use ground-truth labels of the test set for evaluation. We also resize all images to size $64 \times 64$ in order to match the required input size of the GAN. For hyper-parameter tuning, we use the validation loss of corresponding ablations. 

\textbf{Datasets:}  For evaluation on unsupervised few-shot classification, we follow  \cite{hsu2018unsupervised, antoniou2019assume, khodadadeh2018unsupervised}, and evaluate on two  benchmark datasets: Mini-Imagenet \cite{vinyals2016matching} and Tiered-Imagenet \cite{ren2018meta}. 
% For evaluation on K-nearest-neighbor retrieval and K-means clustering, we also follow  work on unsupervised distance-metric learning \cite{iscen2018mining, ye2019unsupervised}, and evaluate on the following benchmark datasets: CUB 200-2011 \cite{wah2011caltech}, Cars196 \cite{krause20133d}, and Stanford Online Product (Product) \cite{song2016deep}. 
Our training is unsupervised, starts from scratch, and does not use other datasets for pre-training.

Mini-Imagenet consists of 100 randomly chosen classes from ILSVRC-2012 \cite{russakovsky2015imagenet}. As in  \cite{donahue2016adversarial, antoniou2019assume,hsu2018unsupervised}, these classes are randomly split into 64, 16, and 20 classes for training, validation, and testing, respectively. Each class has 600 images of size $84\times 84$. Tiered-Imagenet consists of 608 classes of $84 \times 84$ images from ILSVRC-2012 \cite{russakovsky2015imagenet}, grouped into 34 high-level categories. These are divided into 20, 6 and 8 categories for meta-training, meta-validation, and meta-testing. This corresponds to 351, 97 and 160 classes for meta-training, meta-validation, and meta-testing respectively. Tiered-Imagenet minimizes the semantic similarity between the splits compared to Mini-Imagenet. 

% CUB 200-2011 has 11,788 images showing 200 bird classes. We use the standard data split, where 5,864 images of first 100 classes define the training set, and 5,924 images of the remaining classes are the test set. Cars196 has 16,185 images of 196 car classes. We also use the standard data split, where the training set consists of 8,054 images of the first 96 classes and the test set consists of 8,131 images of the remaining car classes. Since CUB 200-2011 and Cars196 do not have a validation split, for tuning hyper-parameters, we further split the standard training dataset into a new training set and validation set. Specifically, we use 70 (30) classes and 70 (26) classes for training (validation) on CUB 200-2011 and Cars196, respectively. Finally, Stanford Online Product (Product) consists of 59,551 training images showing 11,318 object categories, and 60,502 test images showing other 11,316 categories that are not present in the training images. We randomly select 30\% of the Product training images for validation and use the remainder for our unsupervised training.

\textbf{Evaluation metrics:} For few-shot classification,  we first randomly sample $N$ classes from the test classes and $K$ examples for each sampled class, and then classify query images into these $N$ classes. We report the average accuracy over 1000 episodes with $95\%$ confidence intervals of the $N$-way $K$-shot classification.
% specified in Sec.~\ref{sec:testing}. For evaluation of K-nearest-neighbor retrieval and K-means clustering, we use a percentage of the retrieved images that have the same ground-truth class as the query, $R@K$, and the normalized mutual information, NMI, as specified in Sec.~\ref{sec:testing}.

Specifically, we are given support images $x_s$ with labels $y_s \in L_{\text{test}}$ sampled from $N=|L_{\text{test}}|$ classes which have not been seen in training. Each class has $K$ examples, $K\le 5$. Given a query image, $x_q$, we predict a label $\hat{y}_q \in L_{\text{test}}$ of the query as follows. After computing deep representations $z_q=D_z(x_q)$ and $z_s=D_z(x_s)$ of the query and support images, for every class $n=1,\dots,N$, we estimate its mean prototype vector $c_n$ as $
 c_n = \frac{1}{K} \sum_{s, y_s=n}  z_s$, and
take the label of the closest $c_n$ to $z_q$ as our solution:
\begin{equation}
 \hat{y}_q = \hat{n}=\argmin_n \Delta(z_q,c_n),
\label{eq:recognition}
 \end{equation}
where $\Delta$ is a distance function (e.g.,  \eqref{eq:distance}). The same formulation of few-shot recognition is used in \cite{snell2017prototypical}.

\textbf{Implementation details:}
Our implementation is in Pytorch \cite{paszke2017automatic}. Our backbone GAN is the Spectral Norm GAN (SN-GAN) \cite{miyato2018spectral} combined with the self-modulated batch normalization \cite{chen2018self}. All images are resized to $64 \times 64$ pixels, since the SN-GAN  cannot reliably generate higher resolution images beyond $64 \times 64$. There are 4  blocks of layers in both $G$ and $D$. The latent code $z'$ and image representation $z$ have length $d=128$. We use an ADAM optimizer \cite{kingma2014adam} with the learning rate of $5e^{-4}$. We empirically observe convergence of the first and second training stages after $T_1=50000$ and $T_2=10000$ iterations, respectively. $D$ is updated in $\mathcal{T}=3$ iterations for every update of $G$. In all experiments, we set $\gamma = 1, \beta = 1, \lambda = 0.2, \rho=0.5$ as they are empirically found to give the best performance. In the first and the second training stages, the mini-batch size is 128 and 32, respectively.  Our image masking places a $16 \times 16$ patch at $4 \times 4$ locations of the regular grid in the original image, where the patch brightness is equal to the average of image pixels.  It is worth noting that we do not employ other popular data-augmentation techniques in training (e.g., image jittering, random crop, etc.).

\textbf{Ablation study:} The following variants test how individual components of our approach affect performance:  
\begin{itemize}[itemsep=-1pt,topsep=2pt, partopsep=1pt]
    \item T: An architecture that is not a GAN, but consists of only the discriminator network from the SN-GAN \cite{miyato2018spectral}, and  the discriminator's encoding head $D_z$ is trained on the triplet loss only. This variant tests how distance-metric learning affects performance without adversarial learning. 
    \item Gc and Gd: SN-GAN \cite{miyato2018spectral} with continuous uniformly distributed and discrete Bernoulli-distributed elements of the latent code $z'$, respectively.
    \item GcM and GdB: Gc and Gd are extended with the MSE and BCE reconstruction loss, respectively.  
    \item GcT1 and GcT2 = Gc + T: Gc is extended with the triplet loss, and the discriminator is trained in a single stage with total loss $L_D = L_D^{\text{adv}}+ \gamma L_D^{\text{triplet}}$ and in two stages as specified in Alg.~\ref{algo:1}, respectively. These two variants compare performance of single-stage and two-stage training.
    \item GdT1 and GdT2 = Gd + T: Gd is extended similarly as Gc for  GcT1 and GcT2.
    \item GcMT1 and GcMT2 = Gc + MSE + T: Gc is extended with the MSE reconstruction loss and the triplet loss, and the discriminator is trained in a single stage with total loss $L_D = L_D^{\text{adv}}+ \gamma L_D^{\text{mse}}+ \gamma L_D^{\text{triplet}}$ and in two stages as specified in Alg.~\ref{algo:1}, respectively.
    \item GdBT1 and  GdBT2 = Gd + BCE + T: Gd is extended the BCE reconstruction loss and the triplet loss, and the discriminator is trained in a single stage with total loss $L_D = L_D^{\text{adv}}+ \gamma L_D^{\text{bce}}+ \gamma L_D^{\text{triplet}}$ and in two stages as specified in Alg.~\ref{algo:1}, respectively.
\end{itemize}

\begin{table}[b]
\caption{Average accuracy of few-shot image classification on test classes in Mini-Imagenet. GdBT2 with the $16\times16$ masking patch gives the best results.}
\label{table:ablation_study}
{\normalsize
\begin{center}
\begin{tabular}{|l|c|c|}
\hline
\textbf{Ablations} & 1-shot & 5-shot \\
\hline
T & 31.23 $\pm$ 0.46 & 41.91 $\pm$ 0.53 \\
Gc & 34.52 $\pm$ 0.57 & 44.24 $\pm$ 0.72 \\
Gd & 34.84 $\pm$ 0.68 & 44.73 $\pm$ 0.67 \\
GcM & 43.49 $\pm$ 0.76  & 57.62 $\pm$ 0.73 \\
GdB & 43.51 $\pm$ 0.77  & 57.94 $\pm$ 0.76 \\
\hline
GcT1 & 41.95 $\pm$ 0.47 & 50.62 $\pm$ 0.54 \\
GdT1 & 42.13 $\pm$ 0.52 & 51.39 $\pm$ 0.63 \\
GcMT1 & 45.13 $\pm$ 0.63 & 58.87 $\pm$ 0.69 \\
GcBT1 & 45.43 $\pm$ 0.78 & 58.96 $\pm$ 0.72 \\
\hline
GcT2 & 42.36 $\pm$ 0.45 & 52.76 $\pm$ 0.52 \\
GdT2 & 43.13 $\pm$ 0.53 & 53.39 $\pm$ 0.78 \\
GcMT2 & 46.72 $\pm$ 0.73 & 60.92 $\pm$ 0.74 \\
{\bf GdBT2} & \textbf{48.28 $\pm$ 0.77} & \textbf{66.06 $\pm$ 0.70} \\
\hline 
GdBT2 $8\times8$ & 46.23 $\pm$ 0.77 & 60.46 $\pm$ 0.64 \\
{\bf GdBT2 $\bm{16\times16}$} & \textbf{48.28 $\pm$ 0.77} & \textbf{66.06 $\pm$ 0.70} \\
GdBT2 $32\times32$ & 44.40 $\pm$ 0.64 & 57.93 $\pm$ 0.49 \\
\hline
\end{tabular}
\end{center}
}
\end{table}

\begin{table*}
\caption{Unsupervised few-shot classification on Mini-Imagenet and Tiered-Imagenet. At the bottom, we show results of recent fully-supervised approaches  to  few-shot classification as our upper bound.}
\label{table:compare-mini-imagenet}
\centering
{\normalsize
\begin{tabular}{|l|c|c|c|c|}
\hline
& \multicolumn{2}{|c|}{\textbf{Mini-Imagenet, 5-way}} & \multicolumn{2}{|c|}{\textbf{Tiered-Imagenet, 5-way}} \\
\hline
\textbf{Unsupervised Methods} & 1-shot & 5-shot & 1-shot & 5-shot \\
\hline
SN-GAN \cite{miyato2018spectral} & 34.84 $\pm$ 0.68 & 44.73 $\pm$ 0.67  & 35.57 $\pm$ 0.69 & 49.16 $\pm$ 0.70 \\
AutoEncoder \cite{vincent2010stacked} & 28.69 $\pm$ 0.38 & 34.73 $\pm$ 0.63  & 29.57 $\pm$ 0.52 & 38.23 $\pm$ 0.72 \\
Rotation \cite{gidaris2018unsupervised} & 35.54 $\pm$ 0.47 & 45.93 $\pm$ 0.62  & 36.90 $\pm$ 0.54 & 51.23 $\pm$ 0.72 \\
\hline 
BiGAN kNN \cite{donahue2016adversarial} & 25.56 $\pm$ 1.08 & 31.10 $\pm$ 0.63 & - & - \\
AAL-ProtoNets \cite{antoniou2019assume}  & 37.67 $\pm$ 0.39 & 40.29 $\pm$ 0.68 & - & - \\
UMTRA + AutoAugment \cite{khodadadeh2018unsupervised} & 39.93 & 50.73 & - & - \\
CACTUs-ProtoNets \cite{hsu2018unsupervised} & 39.18 $\pm$ 0.71 & 53.36 $\pm$ 0.70  & - & -\\
\hline
% Our GdBT2 & \textbf{47.40 $\pm$ 0.78} & \textbf{61.63 $\pm$ 0.72} & \textbf{47.48 $\pm$ 0.78} & \textbf{64.39 $\pm$ 0.74} \\
Our GdBT2 & \textbf{48.28 $\pm$ 0.77} & \textbf{66.06 $\pm$ 0.70} & \textbf{47.86 $\pm$ 0.79} & \textbf{67.70 $\pm$ 0.75} \\
\hline
\hline
{\bf Fully-supervised  Methods} &&&&\\
ProtoNets \cite{snell2017prototypical} & 46.56 $\pm$ 0.76 & 62.29 $\pm$ 0.71 & 46.52 $\pm$ 0.72 & 66.15 $\pm$ 0.74 \\
% MetaOptNet \cite{Lee_2019_CVPR} & 62.64 $\pm$ 0.62 & 78.63 $\pm$ 0.46 & 65.99 $\pm$ 0.72 & 81.56 $\pm$ 0.53 \\
\hline
\end{tabular}
}
\end{table*}

Table.~\ref{table:ablation_study} presents our ablation study on the tasks of unsupervised 1-shot and 5-shot image classification on Mini-Imagenet. As can be seen, the variants with discrete Bernoulli-distributed elements of $z'$, on average, achieve better performance by $0.5\%$  than their counterparts with continuously distributed latent codes $z'$. Incorporating the reconstruction loss improves performance up to $9\%$ over the variants whose discriminator does not reconstruct the latent codes. While the variant T is the worst, integrating the triplet loss with adversarial learning  outperforms the variants Gc and Gd which use only adversarial learning by more than $8\%$. Relative to  Gc and Gd, larger performance gains are obtained by additionally using the reconstruction loss in GcM and GdB than using the triplet loss in GcT2 and GdT2. Finally, the proposed two-stage training in Alg.~\ref{algo:1} gives better results than a single-stage training, due to, in part, difficulty to optimize hyper parameters. The bottom three variants in Table.~\ref{table:ablation_study} evaluate GdBT2 for different sizes of the masking patch. From Table.~\ref{table:ablation_study}, GdBT2 gives the best results when the masking patch has size $16\times 16$ pixels, and we use this model for comparison with prior work.

\begin{figure*}[ht!]
    \centering
    \includegraphics[scale=0.4]{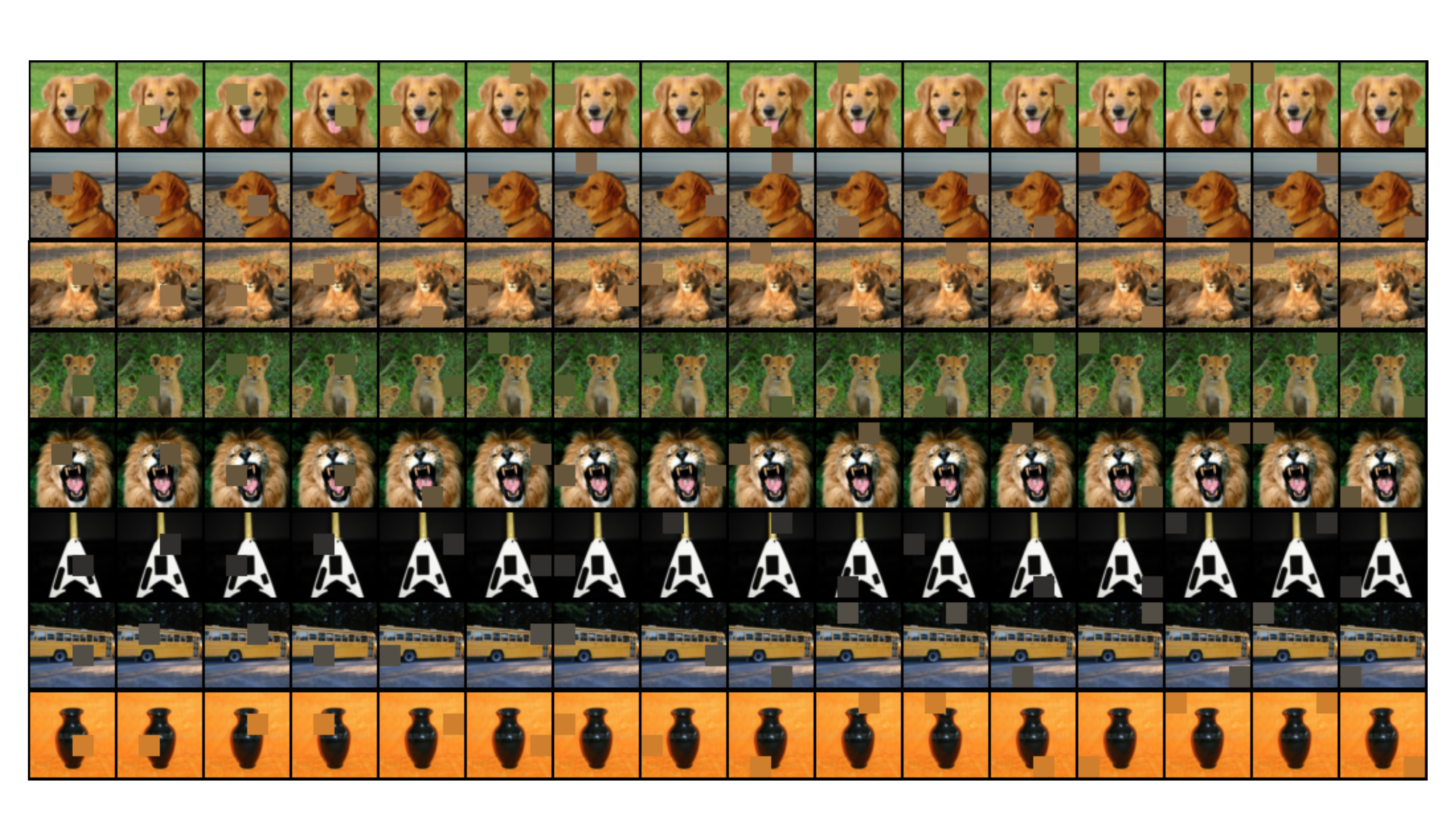}
    \caption{Our image masking with rectangular patches for Mini-Imagenet. In every row, the images are organized from left to right in the descending order by their estimated distance to the original (unmasked) image.}
    \label{fig:qualitative}
\end{figure*}

\textbf{Comparison with state of the art:}
Tab~\ref{table:compare-mini-imagenet}  compares our GdBT2 with the state of the art on the tasks of unsupervised 1-shot and 5-shot image classification on Mini-Imagenet and Tiered-Imagenet. For fair comparison, we follow the standard label assignment to query images as in  \cite{snell2017prototypical}. As can be seen, we significantly outperform the state of the art \cite{hsu2018unsupervised} by $9\%$ in 1-shot and nearly $13\%$ in 5-shot settings of Mini-Imagenet dataset. At the bottom,  Tab.~\ref{table:compare-mini-imagenet} also shows results of recent fully-supervised approaches  to  few-shot classification, as their performance represents our upper bound. Surprisingly, our GdBT2 achieves significantly outperforms the fully supervised ProtoNets \cite{snell2017prototypical}. Our approach's performance can be even further boosted in practice since we can make use of abundant unsupervised data while supervised approaches are not applicable.

\textbf{Qualitative Results}: 
Fig.~\ref{fig:qualitative} illustrates our masking procedure for generating positive and negative images in the triplets for distance-metric learning. In each row, the images are organized from left to right by their estimated distance to the original (unmasked) image in the descending order, where the rightmost image is the closest. From Fig.~\ref{fig:qualitative},  masked images that are the closest to the original have the masking patch in the image corner, and thus are good candidates for the positives in the triplets. Also, when the masking patch covers central image areas the resulting masked images have greater distances from the original, and are good candidates for the negatives in the triplets.

\section{Conclusion}
\label{sec:conclusion}
We have addressed unsupervised few-shot object recognition, where all training images are unlabeled and do not share classes with test images. We have extended the vanilla GAN so as to integrate the standard adversarial learning with our two new strategies for self-supervised learning. The latter is specified by enforcing the GAN's discriminator to: (a) reconstruct the randomly sampled latent codes, and (b) produce image encodings that respect similarity relationships of images. Results of an extensive ablation study on few-shot classification demonstrate that integrating: (i) triplet loss with adversarial learning outperforms the vanilla GAN by more than 8\%, (ii)  reconstruction loss with adversarial learning gives a performance gain of more than 9\%, and (iii) both triplet and reconstruction losses with adversarial learning improves performance by 13\%. In unsupervised few-shot classification, we outperform the state of the art by  $9\%$ on Mini-Imagenet in 1-shot and $13\%$ in 5-shot settings.

\section*{Acknowledgement} This work was supported in part by DARPA XAI Award N66001-17-2-4029 and DARPA MCS Award N66001-19-2-4035.

{\normalsize
\bibliographystyle{IEEEtran}
\bibliography{IEEEfull}
}

\end{document}